\documentclass[11pt]{article}
\usepackage{xspace,epsfig,syntonly,amsmath}

\usepackage[margin=1in]{geometry}

\usepackage{fancyhdr}
\usepackage{wrapfig}
\usepackage{tikz}
\usepackage{amsmath,epsfig,syntonly}
\usepackage{xspace,latexsym,syntonly}
\usepackage{amssymb}
\usepackage{amsfonts}
\usepackage{subfigure}
\usepackage{multirow}
\usetikzlibrary{snakes}
\usepackage{balance}

\usepackage{times,courier,palatino,helvet}
\usepackage[T1]{fontenc}
\RequirePackage{fix-cm}
\usepackage{microtype}

\usepackage{textcomp}

\usepackage{epstopdf}
\usepackage{hyperref}
\usepackage{multirow}
\usepackage{booktabs}
\usepackage{floatrow}
\usepackage{enumitem}
\setlist{nolistsep}

\newcommand{\autoten}{\textsc{AutoTen}\xspace}

\usepackage[font=small,labelfont=bf,skip=-15pt]{caption}
\DeclareCaptionType{copyrightbox}

\usepackage[compact]{titlesec}
\psfull % switch back .ps figures suppressed by draft option

\newcommand{\papertitle}{Balancing Interpretability and Predictive Accuracy for Unsupervised Tensor Mining}

\title{\papertitle}
\author{Ishmam Zabir (\url{izabi001@ucr.edu}) {\it University of California Riverside},\\ Evangelos Papalexakis (\url{epapalex@cs.ucr.edu})  {\it University of California Riverside}}
\date{}

\begin{document}
\maketitle

\begin{abstract}
The PARAFAC tensor decomposition has enjoyed an increasing success in exploratory multi-aspect data mining scenarios. A major challenge remains the estimation of the number of latent factors (i.e., the rank) of the decomposition, which yields high-quality, interpretable results. Previously, we have proposed an automated tensor mining method which leverages a well-known quality heuristic from the field of Chemometrics, the Core Consistency Diagnostic (CORCONDIA), in order to automatically determine the rank for the PARAFAC decomposition. In this work we set out to explore the trade-off between 1) the interpretability/quality of the results (as expressed by CORCONDIA), and 2) the predictive accuracy of the results, in order to further improve the rank estimation quality. Our preliminary results indicate that striking a good balance in that trade-off benefits rank estimation.
\end{abstract}

\section{Extended Summary}

%In exploratory data mining applications, the case is very frequently the following: we are given a piece of (usually very large) data that is of interest to a domain %expert, and we are asked to identify regular and irregular patterns that are potentially useful to the expert who is providing the data.
Very frequently, tensor mining is done in an entirely unsupervised way, since ground truth and labels are either very expensive or hard to obtain. Our problem, thus, is:  given a potentially very large and sparse tensor, and its $R$-component decomposition, compute a quality measure for that decomposition. Subsequently, using that quality metric, we would like to identify a ``good'' number $R$ of components, and ultimately minimize human intervention and trial-and-error fine tuning.

This problem is extremely hard. In fact, even computing the rank of a tensor has been shown to be an NP-hard problem, in stark contrast to the matrix rank which can be easily computed in polynomial time.
Fortunately, there exist heuristics that are able to assist with the above problem and have been shown to work well in practice, in the field of Chemometrics. Such a powerful and intuitive heuristic is the so-called ``Core Consistency Diagnostic'' \cite{bro2003new}, which given a tensor and its PARAFAC decomposition, provides a quality measure, which we can in turn use as a proxy of how interpretable our results are.
%However, this diagnostic has been specifically designed for fully dense and small datasets, and is not able to scale to large and sparse data.
%In \cite{papalexakis2015fastcorcondia} exploiting sparsity, the PI introduced a {\em provably exact algorithm} that operates on at least {\em two orders of %magnitude larger data} than the state of the art, which enabled quality assessment on large real datasets for the first time.
In our previous work \cite{papalexakis2016autoten} we introduce \autoten, a comprehensive unsupervised and automatic tensor mining method that provides quality assessment of the results by trading off the decomposition quality (measured via the Core Consistency Diagnostic), and the number of latent components one can extract with high-enough quality. \autoten outperforms state-of-the-art approaches, such as \cite{zhaobayesian} and \cite{morup2009automatic}.

The PARAFAC decomposition presents a unique opportunity: it admits a very intuitive interpretation (that of soft-clustering the entities involved in all the modes of the tensor) and albeit heuristic, we have a few means of judging how well our results adhere to the PARAFAC model, and in turn, how interpretable they are. On the other hand, the PARAFAC decomposition has been successfully used for collaborative filtering \cite{xiong2010temporal} where the {\em prediction accuracy} is the focus. Thus, PARAFAC is capable of achieving both high interpretability and high prediction accuracy, but there is a catch.
Unfortunately, a highly predictive decomposition need not be of high quality for interpretability purposes, and vice versa: for instance, a degenerate PARAFAC decomposition where two or more components are linearly dependent would yield the same prediction accuracy if only one of those components remained in the results after proper scaling, however, the redundant components hurt the interpretability of our results.

\begin{figure}[!ht]
	\begin{center}
		\includegraphics[width = 1\textwidth]{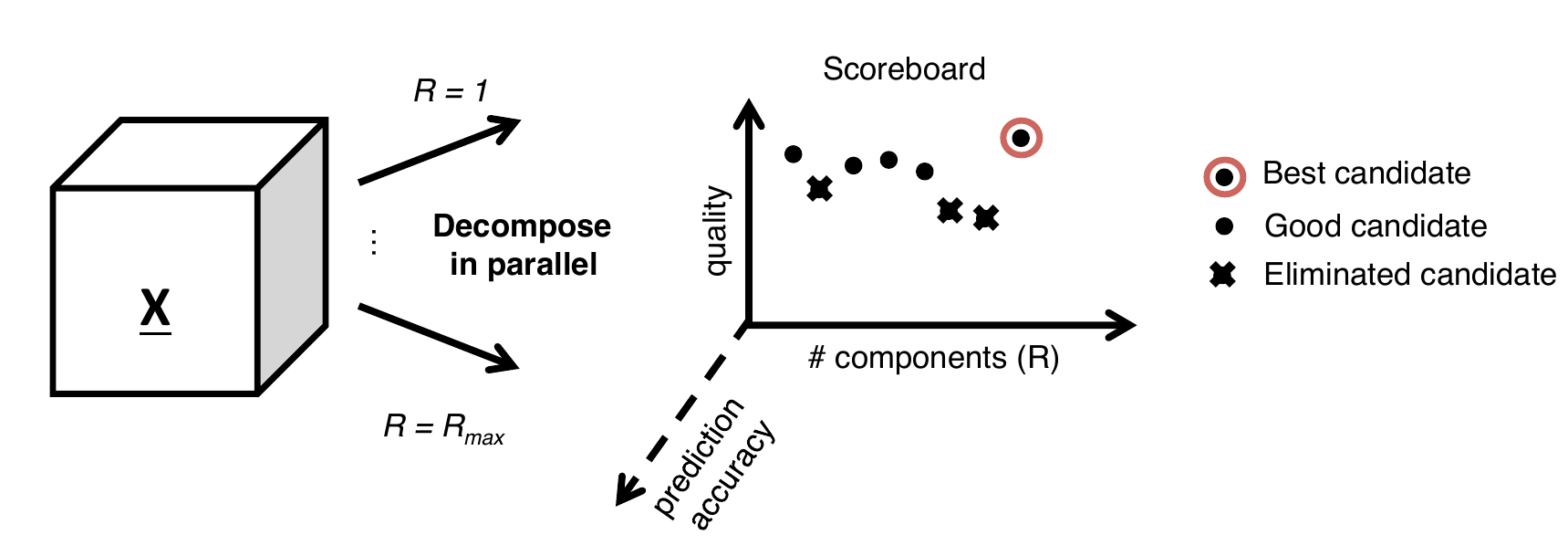}
		\vspace{-0.5in}		
		\caption{Trading off interpretability and prediction accuracy.}
		\label{fig:autoten_tradeoff}
	\end{center}
\end{figure}

Here, we propose to augment \autoten in order to identify a decomposition that combines the best of both worlds: high interpretability and high prediction accuracy. At a high level, as shown in Figure \ref{fig:autoten_tradeoff}, our method will seek to find a set of results that give the best trade-off between quality (which might be multi-dimensional) and as prediction accuracy, which is measured by root mean squared error (RMSE) on held-out data. At the same time, our method will aim at maximizing the number of components we can extract that yield high interpretability {\em and} high prediction accuracy. This is a multi-objective optimization problem and part of our future research revolves around how to best frame it so that we can solve it efficiently without having to exhaustively search the space of solutions. For the sake of our preliminary work here, we use a parameter-free 2-means clustering scheme, similar to \cite{papalexakis2016autoten} which, in a nutshell, separates the different solutions into two clusters, one with high-quality, acceptable solutions, and one with unacceptable solutions, and subsequently identifies a ``good'' solution from the high-quality cluster.

\begin{figure}[!ht]
	\begin{center}
		\includegraphics[width = 0.8\textwidth]{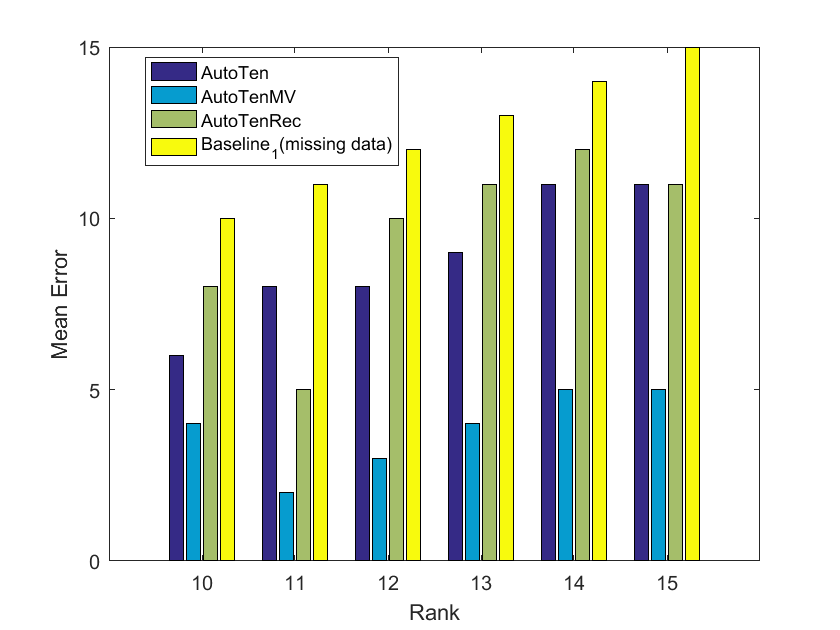}
		\caption{Preliminary rank estimation results.}
		\label{fig:results}
	\end{center}
\end{figure}

For our preliminary analysis we use the N-way toolbox for Matlab implementations of the Frobenius norm ALS algorithm for PARAFAC, as well as the missing value PARAFAC algorithm. We generated $R\times R\times R$ random tensors with rank $R$ recover its rank using the following methods:
\begin{itemize}
	\item {\autoten}: This is the method proposed in \cite{papalexakis2016autoten} which maximizes CORCONDIA and number of components.
	\item {\autoten-REC}: This is the augmented \autoten where both CORCONDIA and reconstruction error are considered as features. In \cite{kamstrup2013core} the authors provide guidelines on using the Core Consistency in conjunction with the fit for rank selection, however, our work is the first method that automatically does so.
	\item {\autoten-MV}: This is the augmented \autoten where both CORCONDIA and missing value prediction are considered as features.
	\item {Baseline-1}:  This method uses the Frobenius norm reconstruction error and returns the rank of the decomposition for which the reconstruction error stopped decreasing (by a small number, set to $10^{-6}$).
\end{itemize}
In Figure \ref{fig:results} we show preliminary results which indicate that balancing interpretability (as expressed through the Core Consistency Diagnostic) and some sort of quality (either reconstruction or predictive accuracy) yields higher-quality rank estimation than the baselines.

%\footnotesize
%\balance
%\newpage
%\setcounter{page}{1}
%\cfoot{E-\thepage}
%\rhead{References}
\bibliographystyle{plain}
\bibliography{BIB/vagelis_refs}

\end{document}